\documentclass[runningheads]{llncs}
\usepackage[T1]{fontenc}
\usepackage{graphicx}
\usepackage{amsmath}
\usepackage{subfigure}
\usepackage{multirow}
\usepackage{url}
\usepackage{fancyhdr}
\usepackage{color}

\usepackage{subcaption}
\usepackage{times}
\usepackage{latexsym}
\usepackage{microtype}
\usepackage{inconsolata}
\usepackage{amsmath,amssymb}
\usepackage{booktabs}
\usepackage{multirow}
\usepackage{graphicx}
\usepackage{enumitem}
\usepackage{xcolor}
\usepackage{hyperref}
\hypersetup{colorlinks=true,linkcolor=blue,citecolor=blue,urlcolor=blue}

\begin{document}
\title{Multi-Granularity Sentiment Integration for LLM-Based Multimodal Sentiment Analysis}
\titlerunning{MGSI: Multi-Granularity Sentiment Integration for MSA}
%
\author{Shanshan Lin\inst{1} \and
Yuesheng Wu\inst{1} \and
Chao Chen\inst{2} \and
Yizhe Yang\inst{1} \and
Zhihao Chen\inst{3} \and \\
Zexian Yang\inst{1} \and 
Xiangwen Liao\inst{1}\thanks{Corresponding author.} 
}
\authorrunning{S. Lin et al.}
\institute{Fuzhou University, Fuzhou, China \and
Harbin Institute of Technology (Shenzhen), Shenzhen, China
\and
Jiangxia University, Fuzhou, China \\
\email{liaoxw@fzu.edu.cn}
}
\maketitle              
\begin{abstract}
Multimodal sentiment analysis (MSA) aims to predict sentiment polarity and intensity from heterogeneous inputs such as text, audio, and vision. 
While large language models (LLMs) offer strong semantic priors for MSA, effectively incorporating audio and visual signals effectively remains challenging. 
A key challenge is that audio and visual sentiment cues evolve over different temporal scales, yet many LLM-based methods compress these signals through shallow projection or coarse pooling before fusing them with text, which can weaken cross-modal alignment and erase fine-grained affective information.
We propose MGSI, a multi-granularity sentiment integration framework for LLM-based MSA. 
MGSI first encodes audio and visual streams at short-, medium-, and long-range temporal scales, preserving both local variations and global affective trends. 
It then refines non-text features through text-guided alignment, and applies polarity- and intensity-aware enhancement to better handle ambiguous and near-neutral samples. 
The resulting multimodal representation is finally compressed into a small set of pseudo-tokens for efficient conditioning of a frozen LLM.
Experiments on four public benchmarks show that MGSI substantially outperforms frozen-LLM baselines and remains competitive with strong multimodal methods. 
Further ablation and sensitivity analyses support the effectiveness of multi-granularity temporal modeling, text-guided refinement, and adaptive sentiment calibration.

\end{abstract}
\section{Introduction}
Multimodal sentiment analysis (MSA) aims to predict sentiment polarity and intensity from heterogeneous signals such as text, audio, and vision.
This task is challenging because affective meaning is often conveyed jointly by lexical content, vocal prosody, facial expressions, and body movements, whose interactions can complement or modify one another \cite{gandhi2023review}.
Thus, MSA requires not only textual semantic understanding, but also temporal modeling of non-text modalities and coordinated cross-modal reasoning.

Recent MSA studies have advanced multimodal modeling through sentiment-aware contrastive learning \cite{yang-etal-2024-clgsi}, dynamic modality weighting \cite{feng2024kuda}, long-range sequence modeling \cite{he2025msamba}, and robustness to missing or imperfect modalities \cite{li2024umdf,li2025tfmamba,huang2026recap}.
Despite these efforts, two issues remain.
First, many methods rely on a single temporal abstraction or compress audio-visual streams too early, obscuring short-, medium-, and long-range affective cues.
Second, heavily summarized multimodal representations may lose fine-grained sentiment dynamics before cross-modal fusion and prediction.

This challenge remains for large language models (LLMs) \cite{qwen2.5,glm2024chatglm}, where strong textual reasoning does not automatically yield effective modeling of temporally evolving non-text cues. 
Audio and visual features are not native linguistic tokens and must therefore be transformed into representations that are semantically aligned with text and structurally compatible with the LLM input space. 
Textualization-based methods, such as DEVA \cite{wu2025deva}, convert visual and acoustic signals into textual emotional descriptions before LLM reasoning.
Adapter-based methods, including MSE-Adapter \cite{yang2025mseadapter}, Mixture of Multimodal Adapters \cite{chen2025mma}, and the Modal Feature Optimization Network with Prompt (MFON) \cite{zhang-etal-2025-modal}, transform non-text features into representations suitable for frozen LLMs.
These recent approaches mainly investigate how multimodal evidence is transformed or injected into an LLM, whereas less attention is paid to preserving sentiment-relevant temporal structure before this transformation.
Thus, the main bottleneck in LLM-based MSA is not only how to map audio-visual features into the LLM embedding space, but also which sentiment-relevant temporal structures should survive this mapping.

To address this issue, we propose \emph{Multi-Granularity Sentiment Integration} (\textbf{MGSI}), an LLM-based MSA framework that treats multimodal adaptation as a structured temporal-semantic compression rather than simple projection. 
The key idea is to preserve sentiment-relevant temporal structure in audio and visual streams \textit{before} compressing them into a compact representation for LLM conditioning. 
Specifically, MGSI first encodes audio and visual inputs with multi-granularity temporal operators to capture local fluctuations, medium-range patterns, and longer-range contextual evolution. 
It then refines non-text representations through text-guided alignment, strengthens polarity discrimination with auxiliary non-neutral supervision, and calibrates ambiguous cases via sentiment intensity modeling. 
Finally, the fused multimodal representation is converted into a small set of pseudo-tokens and injected into a frozen LLM for prediction. 
In this way, MGSI preserves informative multimodal sentiment evidence while retaining the efficiency of lightweight LLM adaptation.

The main contributions of this work are as follows:
\begin{itemize}
    \item We identify temporal-semantic compression as a key issue in LLM-based MSA, where audio-visual signals need to be organized before injection into a frozen LLM.
    \item We propose MGSI, a lightweight adaptation framework that encodes audio and visual streams at multiple temporal granularities before pseudo-token compression.
    \item We introduce text-guided non-text refinement with auxiliary polarity-aware supervision and adaptive residual calibration, making the compressed multimodal representation more sentiment-discriminative.
    \item Experiments on four benchmarks show that MGSI consistently outperforms frozen-LLM prompting and remains competitive with strong multimodal methods. 
\end{itemize}

\section{Related Work}

\subsection{Multimodal Sentiment Representation Learning}
A central problem in multimodal sentiment analysis is to model intra-modal temporal dynamics while capturing cross-modal interactions.
Early studies focused on multimodal fusion, including tensor-based fusion in TFN \cite{zadeh2017tensor}, low-rank fusion in LMF \cite{liu2018lowrank}, and directional cross-modal attention for unaligned sequences in MulT \cite{tsai2019mult}.
Subsequent work improved representation learning through invariant and modality-specific decomposition \cite{hazarika2020misa}, self-supervised modality learning \cite{yu2021selfmm}, 
efficient temporal attention \cite{cheng2021spt}, mutual-information-based fusion \cite{han2021mmim}, and sentiment-aware refinement or supervision \cite{wu2022swrm,qian2023skesl,wu2024multiloss,feng2024kuda,he2025msamba}.

Despite substantial progress, many methods summarize non-text modalities before fusion or rely on a largely uniform temporal abstraction. 
This may miss sentiment evidence expressed through short fluctuations, medium-span patterns, or long-range contextual evolution. 
In contrast, our MGSI explicitly preserves multiple temporal granularities before multimodal compression and LLM adaptation.

\subsection{LLM-Based Multimodal Sentiment Analysis}

Recent studies adapt LLMs to text-centric multimodal sentiment tasks \cite{yang2025llmmsa}.
DEVA \cite{wu2025deva} textualizes visual and acoustic content into emotional descriptions before language-model reasoning, which enables direct use of the LLM's linguistic reasoning ability but changes the original non-text feature interface. 
MSE-Adapter \cite{yang2025mseadapter} introduces a lightweight plug-in that maps multimodal features into a frozen LLM.
Mixture of Multimodal Adapters \cite{chen2025mma} adopts a mixture-of-adapters design, while MFON \cite{zhang-etal-2025-modal} performs prompt-based modal feature optimization.
Together, these methods demonstrate the potential of lightweight LLM adaptation for multimodal sentiment analysis.

Our work is closest to adapter-based LLM approaches.
However, most existing adapters mainly address the compatibility problem between non-text features and LLM input embeddings. In contrast, MGSI focuses on the preceding information organization problem: before compression, audio and visual signals are explicitly structured across multiple temporal granularities and refined with text-guided sentiment cues. 
Because recent methods differ in their non-text input forms and evaluation protocols, we use them primarily for conceptual comparison unless controlled reproduction under the same settings is available.

\subsection{Robust and Adaptive Multimodal Sentiment Learning}
Another line of work studies MSA under imperfect multimodal conditions, such as missing, unreliable, or weakly aligned modalities.
Representative approaches include self-distillation \cite{li2024umdf}, invariant representation learning \cite{zhu2025misr}, cross-lingual disentanglement \cite{chen2025crosslingual}, proxy-driven learning with incomplete data \cite{zhu2025prmf}, text-enhanced sequence modeling \cite{li2025tfmamba}, affective pattern recovery \cite{huang2026recap}, 
and information-bottleneck-based robust learning \cite{huang2026dib}.
These studies show that multimodal performance depends not only on fusion strength, but also on how reliably useful signals are preserved under noise or modality degradation.
Our work shares this motivation but targets a different setting.
Rather than designing a task-specific predictor for incomplete inputs, we study how sentiment-relevant audio-visual information should be temporally and semantically organized before conditioning a frozen LLM.

\section{Method}

\subsection{Task Definition}

Given an utterance with text, audio, and visual modalities, we denote the multimodal input as $\mathcal{X} = (X^{(t)}, X^{(a)}, X^{(v)})$, 
where $X^{(t)}$ is the tokenized text sequence, 
and $X^{(a)}$ and $X^{(v)}$ are the audio and visual feature sequences, respectively.
The goal of multimodal sentiment analysis is to predict a sentiment intensity score $y \in \mathbb{R}$, where positive, zero, and negative values correspond to positive, neutral, and negative sentiment, respectively.

For input preprocessing, 
$E^{(t)}=\mathrm{Embed}(X^{(t)})$ denotes the text embeddings extracted from the input embedding layer of a frozen LLM. 
For each non-text modality $q \in \{a,v\}$, we project the input features into a shared latent space: 
$H_0^{(q)} = X^{(q)} W_0^{(q)} + b_0^{(q)}$.

\begin{figure}
    \centering
    \includegraphics[width=0.85\linewidth]{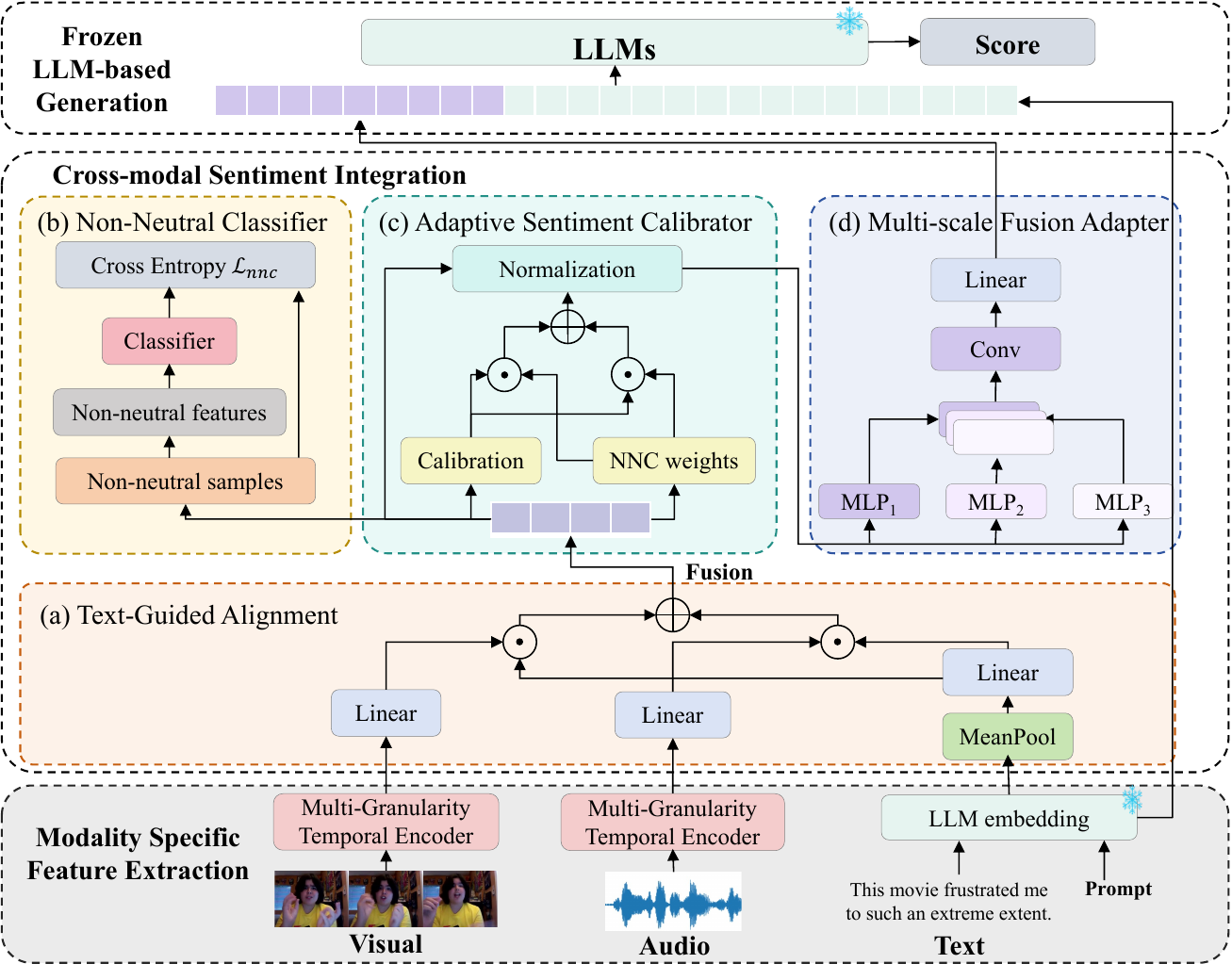}
    \caption{
    Overview of the proposed MGSI framework. 
    Audio and visual streams are first encoded with multi-granularity temporal encoders. Their representations are then refined via (a) text-guided alignment, (b) auxiliary non-neutral classifier, (c) adaptive sentiment calibrator, and (d) multi-scale fusion adapter.
    Finally, the generated pseudo-tokens are injected into the frozen LLM for sentiment prediction. 
    }
    \label{fig:framework}
\end{figure}

\subsection{Overview}

Fig.~\ref{fig:framework} summarizes MGSI, which consists of three stages:
(i) modality specific feature extraction for audio and visual streams (Sec. \ref{sec:mgt}), 
(ii) cross-modal sentiment integration, containing text-guided alignment, non-neutral classifier, adaptive sentiment calibrator, and multi-scale fusion adapter (Secs. \ref{sec:tga}-\ref{sec:msf}), and
(iii) frozen-LLM decoding for sentiment prediction.

MGSI preserves sentiment-relevant temporal structure in non-text modalities \textit{before} compressing them into a compact representation for LLM. Text serves as the semantic anchor, while audio and visual streams offer complementary sentiment evidence. 

\subsection{Multi-Granularity Temporal Encoder (MGT)}
\label{sec:mgt}

Audio and visual sentiment cues may appear at different temporal ranges. 
To capture this temporal structure, we use three parallel branches for each non-text modality.

The \textit{short-term} branch captures local temporal patterns with a 1D convolution:
\begin{equation}
h_s^{(q)} = \mathrm{MaxPool} \bigl(\sigma(\mathrm{Conv1D}(H_0^{(q)}; k_s))\bigr),
\end{equation}
where $k_s$ is the kernel size and $\sigma(\cdot)$ is a nonlinear activation.

The \textit{medium-term} branch enlarges the receptive field with dilated convolution:
\begin{equation}
h_m^{(q)} = \mathrm{MaxPool}\bigl(\sigma(\mathrm{DConv1D}(H_0^{(q)}; r_m^{(q)}))\bigr),
\end{equation}
where $r_m^{(q)}$ is the dilation rate.

The \textit{long-term} branch models longer-range dependencies with a Transformer:
\begin{equation}
\widetilde{H}_l^{(q)} = \mathrm{Transformer}(H_0^{(q)} + Pos^{(q)}),
\end{equation}
where $Pos^{(q)}$ is the positional encoding.
We then apply attention pooling for a summary:
\begin{equation}
\alpha_i^{(q)} =
\frac{\exp(u^\top \widetilde{h}_{l,i}^{(q)})}
{\sum_j \exp(u^\top \widetilde{h}_{l,j}^{(q)})},
\qquad
h_l^{(q)} = \sum_i \alpha_i^{(q)} \widetilde{h}_{l,i}^{(q)},
\end{equation}
where $\widetilde{h}_{l,i}^{(q)}$ is the $i$-th hidden state of the Transformer output and $u$ is a learnable vector.

The three branch summaries are stacked as a three-token sequence and fused by standard multi-head self-attention:
\begin{equation}
\label{eq:hq}
h^{(q)} = \mathrm{MeanPool}\!\left(
\mathrm{MHA}\!\left(
[h_s^{(q)}; h_m^{(q)}; h_l^{(q)}]
\right)\right).
\end{equation}
Here, $[\cdot;\cdot]$ denotes stacking along the token dimension.
$\mathrm{MHA}$ denotes standard multi-head self-attention over the three branch tokens.
$\mathrm{MeanPool}$ averages the three output tokens to obtain the temporally enriched representation $h^{(q)}$.

\subsection{Text-Guided Alignment (TGA)}
\label{sec:tga}

Because text usually provides the clearest semantic anchor in MSA, we use it to refine the audio and visual representations before multimodal fusion.

We first compute a global text summary:
\begin{equation}
g^{(t)}=\mathrm{MeanPool}(E^{(t)}).
\end{equation}
This summary is mapped to a modality-specific gate:
\begin{equation}
r^{(q)}=\sigma(W_g^{(q)} g^{(t)} + b_g^{(q)}), \qquad q \in \{a,v\},
\end{equation}
which modulates the corresponding non-text representation:
\begin{equation}
\hat{h}^{(q)} = h^{(q)} \odot (1 + r^{(q)}),
\end{equation}
where $\odot$ denotes the Hadamard product. 
The aligned audio and visual representations are then fused through a linear projection:
\begin{equation}
h^{(f)} = W_f[\hat{h}^{(a)}; \hat{h}^{(v)}] + b_f.
\end{equation}

\subsection{Non-Neutral Classifier (NNC)}
\label{sec:nnc}

To sharpen polarity discrimination, we introduce an auxiliary binary objective defined over clearly non-neutral samples. 
Let
$\mathcal{I}_{nn} = \{i : |y_i| > \tau\}$
denote the index set of non-neutral instances, where $\tau$ is a threshold on sentiment magnitude. 
A larger $\tau$ restricts supervision to more clearly polarized samples.
For each $i \in \mathcal{I}_{nn}$, the auxiliary label is
\begin{equation}
c_i = \mathbb{I}(y_i > 0),
\end{equation}
where $\mathbb{I}(\cdot)$ is the indicator function.

The classifier predicts $\hat{c}_i$ from $h_i^{(f)}$ using a lightweight MLP:
\begin{equation}
\hat{c}_i = \mathrm{MLP}_{nnc}(h_i^{(f)}).
\end{equation}
The corresponding binary cross-entropy loss is
\begin{equation}
\mathcal{L}_{nnc}
=
-\frac{1}{|\mathcal{I}_{nn}|}
\sum_{i \in \mathcal{I}_{nn}}
\left[c_i \log \hat{c}_i + (1-c_i)\log(1-\hat{c}_i)\right].
\end{equation}

\subsection{Adaptive Sentiment Calibrator (ASC)}
\label{sec:ASC}

Near-neutral or weakly polarized samples are often harder to model because they contain subtler sentiment evidence. 
We therefore introduce an Adaptive Sentiment Calibrator, which performs a sample-adaptive residual correction on the fused feature.
This module adjusts the fused representation with an input-dependent correction coefficient, allowing the model to flexibly refine ambiguous sentiment representations.

Given $h_i^{(f)}$, ASC first predicts a scalar calibration coefficient:
\begin{equation}
\rho_i = \sigma(\mathrm{MLP}_{\rho}(h_i^{(f)})),
\end{equation}
where $\rho_i \in (0,1)$ controls the strength of the correction. The calibrated representation is 
\begin{equation}
\tilde{h}_i =
\mathrm{LN}\!\left(
h_i^{(f)} + \gamma \rho_i \,\mathrm{MLP}_{ASC}(h_i^{(f)})
\right),
\end{equation}
where $\gamma$ is a residual scaling factor and $\mathrm{LN}(\cdot)$ denotes layer normalization.
$\gamma$ is kept small to avoid over-correcting the fused representation.

\subsection{Multi-Scale Fusion Adapter}
\label{sec:msf}

To interface the calibrated multimodal representation with the frozen LLM, we compress it into a small set of pseudo-tokens.
Specifically, we apply $K$ parallel MLP branches to $\tilde{h}$, stack their outputs, and compress the result with a 1D convolution.
The compressed features are then projected into the LLM embedding space to obtain
$P \in \mathbb{R}^{M \times d}$,
where $M$ is the number of pseudo-tokens and $d$ is the hidden size of the frozen LLM.
We set $M=4$ in all experiments, using a small pseudo-token budget to balance representational capacity and the asymptotic attention cost at the LLM interface.

The final input to the frozen LLM is
\begin{equation}
E^{(\mathrm{in})} =
\left[E^{(\mathrm{pre})}; P; E^{(t)}; E^{(\mathrm{suf})}\right],
\end{equation}
where $E^{(\mathrm{pre})}$ and $E^{(\mathrm{suf})}$ denote the prefix and suffix prompt embeddings.

\subsection{Training Objective}

Let $\mathbf{s}=(s_1,\ldots,s_N)$ denote the target token sequence corresponding to the sentiment score. The main autoregressive objective is
\begin{equation}
\mathcal{L}_{gen}
=
-\sum_{n=1}^{N}
\log p(s_n \mid s_{<n}, E^{(\mathrm{in})}).
\end{equation}
The full training objective is
\begin{equation}
\mathcal{L} = \mathcal{L}_{gen} + \lambda_{nnc}\mathcal{L}_{nnc},
\end{equation}
where $\lambda_{nnc}$ controls the weight of the auxiliary non-neutral supervision.

\subsection{Complexity Analysis}

MGSI is designed to preserve multi-scale temporal information while keeping LLM-side computation efficient.
For each non-text modality, the short- and medium-term branches are convolutional and therefore scale linearly with sequence length, while the long-term branch incurs
$\mathcal{O}(L_q^2 d_h + L_q d_h^2)$
due to self-attention.
Because only one lightweight Transformer branch is used per modality, the temporal encoder remains tractable.

The main efficiency gain appears at the LLM interface.
If raw audio and visual sequences were directly injected into the LLM, the self-attention cost would scale as
$\mathcal{O}((L_t + L_a + L_v)^2 d)$.
By compressing non-text features into $M$ pseudo-tokens, MGSI reduces this cost to
$\mathcal{O}((L_t + M)^2 d)$,
where typically $M \ll L_a + L_v$.

In our experiments, $M=4$, so the $L_a+L_v$ non-text feature steps are replaced by four pseudo-tokens before LLM decoding. 
Historical training logs on a single NVIDIA A30 GPU indicate approximately two minutes per training epoch on MOSI.

\section{Experiments}

\subsection{Experimental Settings}

\paragraph{Datasets and metrics}
We evaluate on four public benchmarks: MOSI \cite{zadeh2016multimodal} and MOSEI \cite{bagher-zadeh-etal-2018-multimodal} for English, and SIMS \cite{yu-etal-2020-ch} and SIMS-V2 \cite{liu2022make} for Chinese.
Following prior work \cite{yang2025mseadapter}, we report Acc-2, F1, Acc-7, and MAE on MOSI and MOSEI, and Acc-2, F1, Acc-5, and MAE on SIMS and SIMS-V2.
We follow the official train/validation/test splits of each dataset.
We report the average performance over three random seeds. Due to space limitations, standard deviations are not explicitly reported, while statistical significance is tested using paired t-tests at the 0.05 level.

\paragraph{Backbones and input features}
The frozen LLM backbones are ChatGLM3-6B and Qwen2.5-7B, both used with their official tokenizers.
We use the same pre-extracted audio and visual features across all compared LLM-adapter variants.
The dimensions of text, audio, and visual features are $d=4096$, $d_a=74$, and $d_v=35$, respectively. 

\paragraph{Training protocol}
We train MGSI with AdamW and mixed-precision training.
The maximum number of training epochs is 30 for MOSI, SIMS, and SIMS-V2, and 80 for MOSEI.
The learning rate is initialized to $5\times10^{-5}$.
Linear warmup is applied over the first 10\% of the corresponding total training steps, followed by cosine learning-rate decay. 
Early stopping monitors validation MAE and terminates training after 10 consecutive epochs without improvement.
The batch size is 8.
In addition, we set $\tau=0.1$, $k_s^{(a)}=3$, $r_\mathrm{mid}^{(a)}=2$, $k_s^{(v)}=3$, $r_\mathrm{mid}^{(v)}=2$, $\gamma=0.05$, and $\lambda_{nnc}=0.05$.
All experiments are implemented in Python 3.10 and PyTorch 2.1.0, and run on a single NVIDIA A30 GPU.

\paragraph{Prompting and decoding}
Prompt for English datasets is:
\textit{``Please predict the sentiment intensity of the above multimodal content in the range [-3.0, 3.0]. Response: The sentiment is''}.
For Chinese datasets, we use the corresponding Chinese instruction with the range [-1.0, 1.0].
At inference time, 
we use greedy decoding (temperature 0) and allow up to four generated tokens.

\subsection{Compared Methods}

We compare \textbf{MGSI} against representative multimodal sentiment analysis baselines spanning several model families. 
(i) Tensor-based fusion models: TFN \cite{zadeh2017tensor} and LMF \cite{liu2018lowrank}. 
(ii) Cross-modal interaction and structured representation models: MulT \cite{tsai2019mult}, MISA \cite{hazarika2020misa}, Self-MM \cite{yu2021selfmm}, MMIM \cite{han2021mmim}, CENet \cite{wang2023cenet} and TETFN \cite{wang2023tetfn}. 
(iii) LLM-based approaches: UniMSE \cite{hu2022unimse} and MSE-Adapter \cite{yang2025mseadapter}. 
(iv) In addition, we report direct prompting results of frozen ChatGLM3-6B and Qwen2.5-7B as text-only references.
Classical MSA baselines are reproduced following the same splits, while MSE-Adapter is trained under the same LLM backbone and feature settings for fair comparison.

\subsection{Main Results}

\begin{table*}[t]
\centering
\caption{
Results on \textbf{MOSI} and \textbf{MOSEI}. Best results are in bold. 
$\ast$ and $\dagger$ indicate statistical significance at the 0.05 level compared with the strongest non-LLM method and MSE-Adapter baseline under the same LLM backbone, respectively.}
\label{tab:main_mosi_mosei}
\small
\begin{tabular}{lcccccccc}
\toprule
\multirow{2}{*}{Method} & \multicolumn{4}{c}{MOSI} & \multicolumn{4}{c}{MOSEI} \\
\cmidrule(lr){2-5}\cmidrule(lr){6-9}
 & Acc-2$\uparrow$ & F1$\uparrow$ & Acc-7$\uparrow$ & MAE$\downarrow$ & Acc-2$\uparrow$ & F1$\uparrow$ & Acc-7$\uparrow$ & MAE$\downarrow$ \\
\midrule
TFN & 79.08 & 79.11 & 34.46 & 0.947 & 81.89 & 81.74 & 51.60 & 0.573 \\
LMF & 79.18 & 79.15 & 33.82 & 0.950 & 84.48 & 83.36 & 51.59 & 0.576 \\
MulT & 80.98 & 80.95 & 36.91 & 0.880 & 84.63 & 84.52 & 52.84 & 0.559 \\
MISA & 83.54 & 83.58 & 41.37 & 0.777 & 84.67 & 84.66 & 52.05 & 0.558 \\
CENet & 85.21 & 85.22 & 44.90 & 0.725 & 86.38 & 86.32 & 54.26 & 0.526 \\
Self-MM & 85.46 & 85.43 & 46.67 & 0.708 & 85.15 & 84.90 & 53.87 & 0.531 \\
MMIM & 86.06 & 85.98 & 46.65 & 0.700 & 85.97 & 85.94 & 54.24 & 0.526 \\
UniMSE & 86.90 & 86.42 & 48.68 & 0.691 & \textbf{87.50} & \textbf{87.46} & 54.39 & 0.523 \\
\midrule
Qwen2.5-7B & 82.01 & 81.08 & 38.05 & 0.844 & 72.10 & 72.36 & 35.80 & 0.754 \\
+ MSE-Adapter & 87.17 & 87.07 & 47.03 & 0.647 & 75.97 & 76.27 & 53.66 & 0.532 \\
+ MGSI & 87.80$^\ast$ & 87.77$^{\ast,\dagger}$ & 46.94 & 0.627$^\ast$ & 78.95 & 79.30 & 52.67 & 0.529 \\
\midrule
ChatGLM3-6B & 63.72 & 54.56 & 31.20 & 1.009 & 55.26 & 52.13 & 39.82 & 0.788 \\
+ MSE-Adapter & 88.63 & 88.54 & 46.91 & 0.643 & 75.19 & 75.18 & 54.44 & 0.516 \\
+ MGSI & \textbf{89.60}$^{\ast,\dagger}$ & \textbf{89.57}$^{\ast,\dagger}$ & \textbf{49.15}$^\ast$ & \textbf{0.603}$^\ast$ & 82.72 & 83.00 & \textbf{54.69}$^\ast$ & \textbf{0.509} \\
\bottomrule
\end{tabular}
\end{table*}

\begin{table*}[t]
\centering
\caption{
Results on \textbf{SIMS} and \textbf{SIMS-V2}. Best results are in bold. 
$\ast$ and $\dagger$ indicate statistical significance at the 0.05 level compared with the strongest non-LLM method and MSE-Adapter under the same LLM backbone, respectively.}
\label{tab:main_sims}
\small
\begin{tabular}{lcccccccc}
\toprule
\multirow{2}{*}{Method} & \multicolumn{4}{c}{SIMS} & \multicolumn{4}{c}{SIMS-V2} \\
\cmidrule(lr){2-5}\cmidrule(lr){6-9}
 & Acc-2$\uparrow$ & F1$\uparrow$ & Acc-5$\uparrow$ & MAE$\downarrow$ & Acc-2$\uparrow$ & F1$\uparrow$ & Acc-5$\uparrow$ & MAE$\downarrow$ \\
\midrule
TFN & 78.38 & 78.62 & 39.30 & 0.432 & 80.14 & 80.14 & 52.55 & 0.303 \\
LMF & 77.77 & 77.88 & 40.53 & 0.441 & 74.18 & 73.88 & 47.79 & 0.367 \\
MulT & 78.56 & 79.66 & 37.94 & 0.453 & 80.68 & 80.73 & 54.81 & 0.291 \\
Self-MM & 80.04 & 80.44 & 41.53 & 0.425 & 79.69 & 79.76 & 52.77 & 0.311 \\
TETFN & 81.18 & 80.24 & 41.79 & 0.420 & 79.73 & 79.81 & 54.47 & 0.310 \\
CENet & 77.90 & 77.53 & 33.92 & 0.471 & 79.56 & 79.63 & 53.04 & 0.310 \\
\midrule
Qwen2.5-7B & 76.59 & 77.23 & 41.14 & 0.474 & 78.63 & 78.72 & 41.59 & 0.396 \\
+ MSE-Adapter & 79.26 & 77.22 & 44.90 & 0.393 & 80.81 & 80.59 & 55.57 & 0.293 \\
+ MGSI & 81.93$^{\dagger}$ & 81.26$^{\ast,\dagger}$ & \textbf{48.14}$^{\ast,\dagger}$ & 0.377$^{\ast,\dagger}$ & 81.53$^\ast$ & 81.34$^{\ast,\dagger}$ & 59.26$^{\ast,\dagger}$ & 0.286$^\ast$ \\
\midrule
ChatGLM3-6B & 75.93 & 72.63 & 33.05 & 0.471 & 73.31 & 67.70 & 44.49 & 0.370 \\
+ MSE-Adapter & 79.30 & 76.95 & 42.23 & 0.390 & 81.39 & 81.10 & 52.49 & 0.304 \\
+ MGSI & \textbf{82.67}$^{\ast,\dagger}$ & \textbf{81.81}$^{\ast,\dagger}$ & 46.92$^{\ast,\dagger}$ & \textbf{0.359}$^{\ast,\dagger}$ & \textbf{83.37}$^{\ast,\dagger}$ & \textbf{83.23}$^{\ast,\dagger}$ & \textbf{61.49}$^{\ast,\dagger}$ & \textbf{0.267}$^{\ast,\dagger}$ \\
\bottomrule
\end{tabular}
\end{table*}

Tables~\ref{tab:main_mosi_mosei} and~\ref{tab:main_sims} show that MGSI improves most metrics over frozen-LLM baselines on all four datasets. 
The gains are particularly large for ChatGLM3-6B, where MGSI improves Acc-2 by 25.88, 27.46, 6.74, and 10.06 points on MOSI, MOSEI, SIMS, and SIMS-V2, respectively, while reducing MAE by 0.406, 0.279, 0.112, and 0.103 relative to direct prompting. 
This suggests that direct prompting of frozen LLMs
is insufficient for exploiting non-text sentiment evidence.
MGSI also outperforms MSE-Adapter under both backbones, and remains competitive against other strong multimodal baselines. 
These results suggest that the gains arise not only from adapter-based conditioning, but also from multi-granularity temporal encoding and pre-LLM refinement.

\subsection{Ablation Studies}

\begin{table}[t]
\centering
\caption{
Ablation results using the frozen ChatGLM3-6B backbone on MOSI and SIMS. 
The upper block evaluates the main components, and the lower block analyzes the temporal branches in the MGT encoder.
}
\label{tab:ablation_all}
\small
\begin{tabular}{lcccccccc}
\toprule
Variant & \multicolumn{4}{c}{MOSI} & \multicolumn{4}{c}{SIMS} \\
\cmidrule(lr){2-5}\cmidrule(lr){6-9} 
& Acc-2$\uparrow$ & F1$\uparrow$ & Acc-7$\uparrow$ & MAE$\downarrow$
& Acc-2$\uparrow$ & F1$\uparrow$ & Acc-5$\uparrow$ & MAE$\downarrow$ \\
\midrule
\multicolumn{5}{l}{\textit{Core components}} \\
MGSI        & \textbf{89.60} & \textbf{89.57} & \textbf{49.15} & \textbf{0.603} & \textbf{82.67} & \textbf{81.81} & 46.92 & 0.359 \\
w/o MGT     & 89.30 & 89.28 & 48.42 & 0.617 & 80.48 & 79.56 & 46.08 & \textbf{0.357} \\
w/o NNC     & 89.54 & 89.49 & 48.78 & 0.613 & 82.05 & 80.89 & \textbf{47.13} & 0.361 \\
w/o ASC     & 88.78 & 88.71 & 49.10 & 0.614 & 81.66 & 79.89 & 43.76 & 0.382 \\
\midrule
\multicolumn{5}{l}{\textit{Temporal branches in MGT}} \\
MGSI        & \textbf{89.60} & \textbf{89.57} & \textbf{49.15} & \textbf{0.603} & \textbf{82.67} & \textbf{81.81} & \textbf{46.92} & \textbf{0.359} \\
w/o Short   & 89.48 & 89.41 & 48.80 & 0.608 & 81.49 & 80.08 & 44.16 & 0.370 \\
w/o Mid     & 88.84 & 88.73 & 48.92 & 0.624 & 82.20 & 81.70 & 45.80 & 0.371 \\
w/o Long    & 89.30 & 89.23 & 49.10 & 0.613 & 81.53 & 80.04 & 43.59 & 0.382\\
\bottomrule
\end{tabular}
\end{table}

We conduct ablations on MOSI and SIMS to examine the contribution of each module and each temporal branch.
Table~\ref{tab:ablation_all} shows that the full model achieves the strongest overall balance across metrics.
Removing any core component substantially degrades MOSI and generally also weakens SIMS, indicating that the gains do not come from a single module alone. 
On MOSI, removing ASC leads to the largest overall drop, suggesting that sample-adaptive residual refinement is helpful for stabilizing sentiment representations.
On SIMS, MGT contributes most to Acc-2 and F1, highlighting the importance of explicit multi-granularity temporal modeling. 
NNC provides smaller but consistent gains, which supports its role as auxiliary polarity-aware supervision.

As for temporal branches in MGT, 
removing the mid-term branch causes the largest degradation on MOSI, 
while the short-term branch affects Acc-2 most, and the long-term branch has larger influence on F1 and MAE on SIMS.
Thus, no single temporal scale is uniformly optimal, supporting the design of the multi-granularity temporal encoder.

\subsection{Sensitivity and Design Analysis}

\begin{figure*}[t]
    \centering
    \begin{minipage}[t]{0.32\textwidth}
        \centering
        \includegraphics[width=\linewidth]{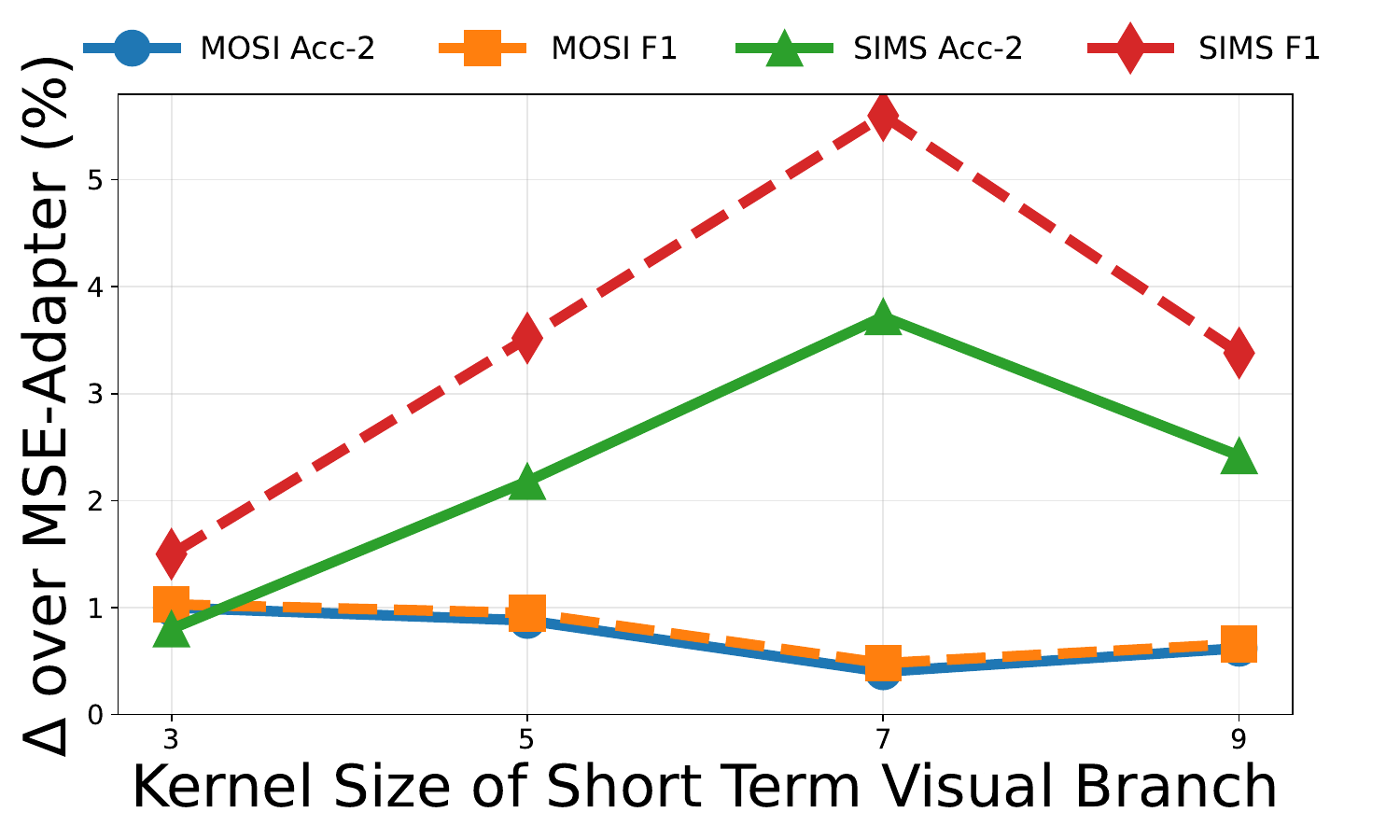}
    \end{minipage}
    \hfill
    \begin{minipage}[t]{0.32\textwidth}
        \centering
        \includegraphics[width=\linewidth]{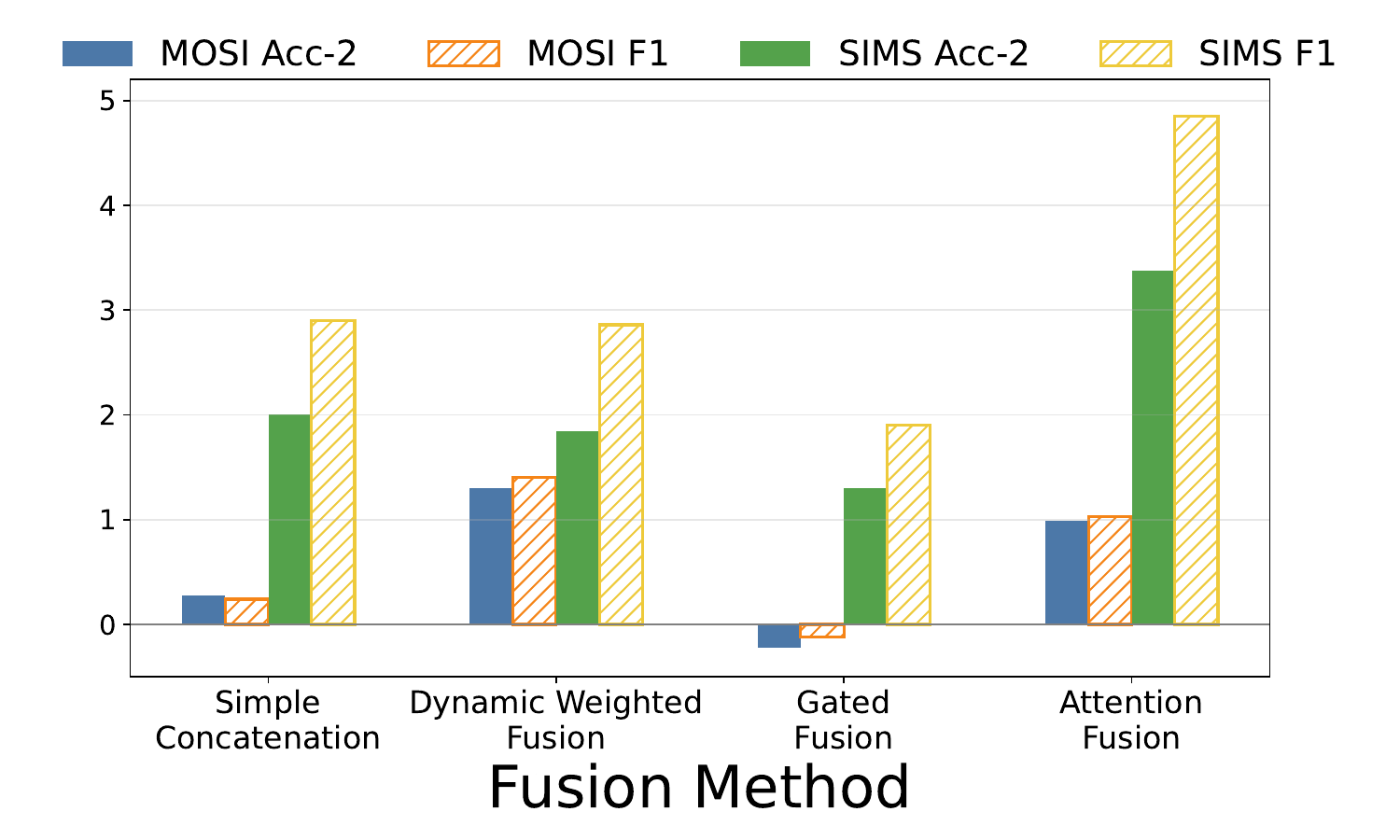}
    \end{minipage}
    \hfill
    \begin{minipage}[t]{0.32\textwidth}
        \centering
        \includegraphics[width=\linewidth]{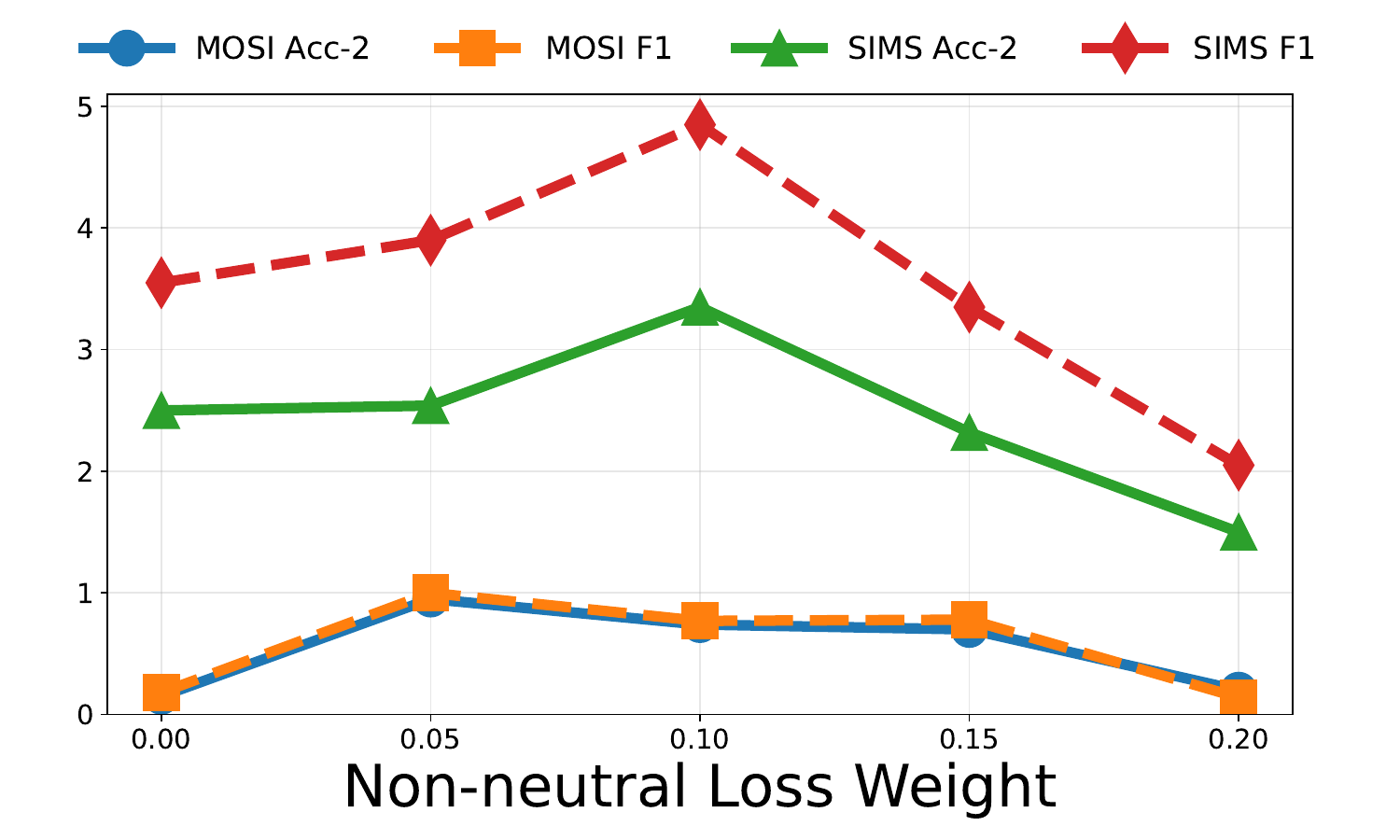}
    \end{minipage}
    \caption{
    Sensitivity and design analysis on MOSI and SIMS. 
    All values denote absolute improvements in percentage points over MSE-Adapter under the same backbone.
    From left to right: sensitivity to the visual short-term kernel size, comparison of branch-fusion strategies, and effect of the non-neutral loss weight $\lambda_{nnc}$.
    }
    \label{fig:analysis_all}
\end{figure*}

Figure~\ref{fig:analysis_all} summarizes the sensitivity and design analysis on MOSI and SIMS. 
All relative changes in the figure are computed with respect to MSE-Adapter under the same backbone. 
For the \textit{short-term visual branch}, the preferred kernel size differs across datasets: MOSI performs best with a kernel of size $3$, whereas SIMS peaks at size $7$. This suggests that the appropriate local receptive field is dataset-dependent. MOSI tends to benefit from more localized visual cues, while SIMS gains from a moderately wider context.

We further compare four \textit{branch-fusion strategies}: simple concatenation, dynamic weighted fusion, gated fusion, and attention-based fusion. 
On MOSI, dynamic weighted fusion achieves the largest gains, improving Acc-2 and F1 by 1.31 and 1.38 points, respectively. 
On SIMS, attention-based fusion performs best, with gains of 3.37 and 4.87 points.
We therefore adopt attention-based fusion as the default setting, because it better captures cross-branch complementarity when temporal patterns are diverse.

Finally, we examine the effect of the \textit{non-neutral loss weight} $\lambda_{nnc}$. 
A moderate value generally provides a better trade-off, although the optimal setting varies by dataset.
When $\lambda_{nnc}$ is too small, the auxiliary supervision is insufficient to sharpen polarity discrimination; when it is too large, the model overemphasizes binary polarity and degrades fine-grained intensity modeling. This result supports the role of NNC as an auxiliary constraint rather than a dominant objective.

\section{Conclusion}

We presented \textbf{MGSI}, an LLM-based multimodal sentiment analysis framework that preserves sentiment-relevant temporal structure in audio and visual streams before conditioning a frozen LLM. 
Instead of directly projecting long non-text sequences into the language model, MGSI performs multi-granularity temporal encoding, text-guided refinement, and compact pseudo-token adaptation. 
This design enables the model to retain richer multimodal affective information while keeping LLM-side computation efficient.
Experiments on four public benchmarks show that MGSI consistently improves frozen-LLM baselines and remains competitive with strong multimodal methods. 
Ablation and sensitivity analyses further indicate that the gains come from the complementary effects of multi-granularity temporal modeling, polarity-aware auxiliary supervision, and adaptive sentiment calibrator, rather than from any single component alone. 

This study also has several limitations.
The framework 
relies on pre-extracted audio and visual features rather than end-to-end raw-signal encoders. In addition, compressing multimodal information into a small number of pseudo-tokens introduces an efficiency-fidelity trade-off, which may become more pronounced for long or highly expressive inputs. Future work may therefore explore adaptive temporal-scale selection, uncertainty-aware calibration, and dynamic pseudo-token allocation. 
Overall, our findings highlight the importance of structured temporal abstraction for effective and efficient multimodal adaptation of frozen LLMs.

%
%
%

\begin{credits}
\subsubsection{\ackname} 
This work was supported by the National Natural Science Foundation of China (No. 62476060). 
\end{credits}

\end{document}